\documentclass{article}
\usepackage[numbers]{natbib}
\usepackage{amsmath}
\usepackage{amsfonts}
\usepackage{amssymb}
\usepackage{multirow}
\usepackage{textcomp}
\DeclareMathAlphabet{\mathpzc}{OT1}{pzc}{m}{it}
\usepackage{fullpage}

\usepackage{ifpdf}
\ifpdf
 	\usepackage[pdftex]{graphicx}
\else
	\usepackage{graphicx}
\fi
\usepackage{subfigure}

  \renewenvironment{thebibliography}[1]{%
    \begin{oldthebibliography}{#1}%
      \setlength{\parskip}{0.7ex}%
      \setlength{\itemsep}{0ex}%
}%
  {%
 \end{oldthebibliography}%
  }
\renewcommand{\refname}{\textbf{\normalsize References}}
\def\bibfont{\small}

\newcommand{\Expon}[1]{\mathrm{Expon}\left( #1 \right )}
\newcommand{\Conj}[1]{\mathrm{Conj}\left(#1 \right)}
\newcommand{\bb}[1]{\boldsymbol{#1}}

\newcommand{\X}{\mathbf{X}}
\newcommand{\V}{\mathbf{V}}
\newcommand{\T}{\bb{\Theta}}

\newenvironment{myindentpar}[1]%
{\begin{list}{}%
         {\setlength{\leftmargin}{#1}}%
         \item[]%
}
{\end{list}}

\usepackage[accepted]{icml2012}
\icmltitlerunning{Bayesian and L$_\mathbf{1}$ Approaches for  Sparse Unsupervised Learning}

\begin{document}

\twocolumn[
\icmltitle{Bayesian and L$_\mathbf{1}$ Approaches for  Sparse Unsupervised Learning}
\icmlauthor{Shakir Mohamed}{shakirm@cs.ubc.ca}
\icmladdress{Department of Computer Science, University of British Columbia}
\icmlauthor{Katherine A. Heller}{kheller@mit.edu}
\icmladdress{Department of Brain and Cognitive Sciences, Massachusetts Institute of Technology}
\icmlauthor{Zoubin Ghahramani}{zoubin@eng.cam.ac.uk}
\icmladdress{Department of Engineering, University of Cambridge}
\vskip0.15in 
]

\begin{abstract}
  The use of $L_1$ regularisation for sparse learning has generated
  immense research interest, with many successful applications in
  diverse areas such as signal acquisition, image coding, genomics and
  collaborative filtering. While existing work highlights the many
  advantages of $L_1$ methods, in this paper we find that $L_1$
  regularisation often dramatically under-performs in terms of
  predictive performance when compared to other methods for inferring sparsity. We
  focus on unsupervised latent variable models, and develop $L_1$
  minimising factor models, Bayesian variants of ``$L_1$'', and
  Bayesian models with a stronger $L_0$-like sparsity induced through
  spike-and-slab distributions. These spike-and-slab Bayesian factor
  models encourage sparsity while accounting for uncertainty in a
  principled manner, and avoid unnecessary shrinkage of non-zero
  values.  We demonstrate on a number of data sets that in practice
  spike-and-slab Bayesian methods outperform $L_1$ minimisation, even
  on a computational budget. We thus highlight the need to re-assess
  the wide use of $L_1$ methods in sparsity-reliant applications,
  particularly when we care about generalising to previously unseen
  data, and provide an alternative that, over many varying conditions,
  provides improved generalisation performance.
\end{abstract}
\section{Introduction}
Over the last decade, there has been tremendous excitement in learning 
parsimonious models using sparsity. Sparse learning is now a significant 
research topic -- this significance being tied to the 
theoretical and practical advancement of sparse learning methods using the 
$L_1$ norm. The use of the $L_1$ norm in penalised regression problems such as 
the Lasso \cite{lasso96}, in natural scene understanding and image coding 
problems \cite{olshausen1996}, and more recently in compressed sensing 
\cite{candes06}, has served to cement the importance and efficacy of the 
$L_1$ norm as a means of inducing sparsity. Among its important properties, 
the $L_1$ norm is the closest convex norm to the $L_0$ norm, has a number of 
provable properties relating to the optimality of solutions and oracle 
properties \cite{vandegeer2009}, and allows for the wide array of tools from 
convex optimisation to be used in computing sparse solutions. With the use of 
sparse methods in increasingly diverse application domains, it is timely to 
now contextualise the use of the $L_1$ norm and critically evaluate its 
behaviour in relation to other competing methods. 
\\ \\
To achieve sparsity, the  idealised but intractable sparsity criterion uses the $L_0$ norm to 
penalise the number of non-zero parameters. To more closely match the $L_0$ objective function, 
we develop here the use of discrete mixture priors for sparse learning, commonly 
referred to as spike-and-slab priors \cite{mitchellSpikeSlab, ishwaran2005}. A 
spike-and-slab is a discrete mixture of a point mass at zero (the spike) and 
any other continuous distribution (the slab). It is  
is similar to the $L_0$ norm in that it imposes a penalty on the number of 
non-zero parameters in a model. We show that spike-and-slab
distributions provide improvements in learning, and that both Bayesian methods 
and the use of the spike-and-slab distribution deserve more prominent 
attention in the vast literature for sparse modelling. 
\\ \\
Our analysis focuses on unsupervised linear latent variable models
(also known as matrix completion models), a class of 
models that are amongst the core tools in the machine learning practitioner's 
toolbox. Factor analysis, the inspiration for this class of models, describes 
real-valued data by a set of underlying factors that are linearly combined to 
explain the observed data. This base model allows for many adaptations, such 
as generalisations to non-Gaussian data \cite{collins02, mohamed09}, or in 
learning sparse underlying factors \cite{dueck04, lee09sepca, 
carvalho2008spike}. In unsupervised learning, a sparse representation is 
desirable in situations where: 1) there are many underlying factors that 
could explain the data, 2) only a subset of which explain the data, and 3) the 
subset is different for each observation. 
\\ \\
After introducing our framework for unsupervised models (section 
\ref{sect:unsupModels}), we develop approaches for sparse Bayesian learning, culminating in a thorough comparative analysis.
Our contributions include:
\vspace{-3mm}
\begin{list}{\labelitemi}{\leftmargin=1em}
	\setlength{\itemsep}{1pt}
	\setlength{\parskip}{0pt}
	\setlength{\parsep}{0pt}
	\item We introduce new generalised latent variable models with strong sparsity, 
providing an important new class of sparse models that can readily  handle 
non-Gaussian and heterogeneous data sets (sect. \ref{sect:l1sparsity}).
	\item We develop a spike-and-slab model for sparse unsupervised 
learning and derive a full MCMC algorithm for it. This MCMC method is
applicable to other models based on discrete-continuous
mixtures and is more efficient than naive samplers (sect.
\ref{sect:spikeslab}). 
	\item We present the first comparison of approaches for sparse unsupervised learning
          based on optimisation methods, Bayesian methods using continuous sparsity-favouring 
priors, and Bayesian methods using the spike-and-slab. We bring these 
methods together and compare their performance in a controlled manner on 
both benchmark and real world data sets across a breadth of model types 
(sect. \ref{sect:experiments}).
\item Interestingly, our results show that strong sparsity in the from
  of spike-and-slab models can outperform the commonly used $L_1$
  methods in unsupervised modelling tasks.
\end{list}
\begin{figure}[t]
\centering
\includegraphics[scale = 0.50]{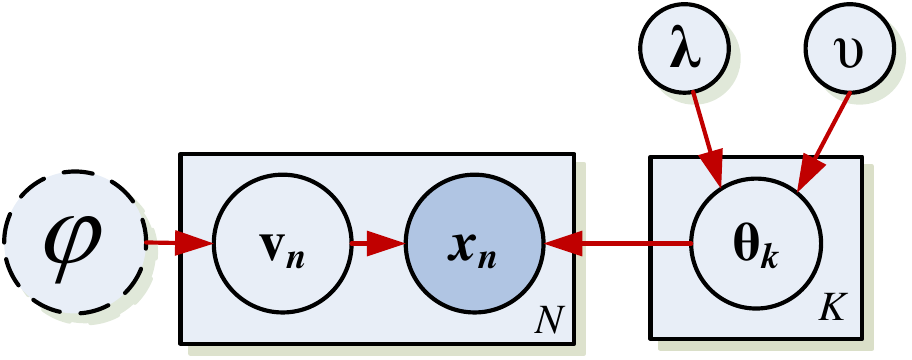}
\caption{Graphical representation for generalised latent variable models.}
\label{fig:graphMod}
\vspace{-2mm}
\end{figure}

\vspace{-4mm}
\section{Unsupervised Latent Variable Models and Sparsity}
\label{sect:unsupModels}
We are concerned with models of the form:
\begin{equation}
\label{eq:basicFactorModel}
 \X = \V \T + \mathbf{E}, \qquad \mathbf{e}_n \sim 
\mathcal{N}(\mathbf{0},\bb\Sigma),
\end{equation}
which is the matrix factorisation problem in which we search for a set
of underlying factors $\V$ and weights $\T$ that are combined to explain the
observed data $\X$. We often consider Gaussian latent variables and Gaussian  
noise with diagonal or isotropic covariances, in which case this model recovers 
the familiar factor analysis and principal components analysis models, 
respectively. If $\V$ is sparse then subsets of the underlying factors 
explain the data and different subsets explain each observed data 
point. 
\\\ \\
Increasingly we do not deal with real-data, which is well described by a 
Gaussian distribution, but data that may be binary, categorical, non-negative 
or a heterogeneous set of these. It is interesting to then consider 
generalisations of the basic model \eqref{eq:basicFactorModel} in which the 
conditional probability of the observed data is defined using the 
exponential family of distributions, as:
\vspace{-2mm}
\begin{eqnarray}
\label{eq:exponSpec}
\mathbf{x}_n | \mathbf{v}_n, \bb{\Theta} \sim \Expon{ \sum_k 
v_{nk}\bb{\theta}_k }; \,  \bb{\theta}_k  \sim \Conj{\bb{\lambda},\nu}
\end{eqnarray}
We use the shorthand $\mathbf{x}_n \sim \Expon{\bb{\psi}}$ to  represent the
exponential family of distributions with natural parameters  $\bb{\psi} =
\mathbf{v}_n\bb{\Theta}$. For this model, the  natural parameters are a sum of
the parameters $\bb{\theta}_k$, weighted by  $v_{nk}$, the points in the latent
subspace corresponding to data point  $\mathbf{x}_n$. For the exponential 
family of distributions, the conditional  probability of $\mathbf{x}_n$ given 
parameter vector $\bb{\psi}$ takes  the form: $ p(\mathbf{x}_n | \bb{\psi})\! =\!
h(\mathbf{x}_n) \exp \left(  s(\mathbf{x}_n)^\top \bb{\psi}   - A(\bb{\psi})
\right) $, where $s(\mathbf{x}_n)$ are the sufficient statistics, $\bb{\psi}$ 
is a vector  of natural parameters and $A(\bb{\psi})$ is the log-partition
function. Probability distributions  that belong to the exponential family also
have natural conjugate prior  distributions, which we use to model the
distribution of the parameters  $\bb{\Theta}$. We use the notation:
$\Conj{\bb{\lambda},\nu}$ as  shorthand for the conjugate distribution, which
has the form: $p(  \bb{\theta}_k) \propto \exp 
(\bb{\lambda}^{\top}\bb{\theta_k} - \nu  A(\bb{\theta}_k))$, with 
hyperparameters $\bb{\lambda}$ and $\nu$, and  $A(\bb{\theta}_k)$ is the same 
log-partition function from the  likelihood function. 
\\ \\ 
Figure \ref{fig:graphMod} is a graphical representation of general 
unsupervised models; the shaded node $\mathbf{x}_n$ represents the 
observed data item $n$. The plate notation represents replication of variables 
and the dashed node $\bb{\varphi}$ represents any appropriate prior 
distribution for the latent variables $\mathbf{v}_n$. The observed data forms 
an $N \times D$ matrix $\X$, with rows $\mathbf{x}_n$. $N$ is the number of 
data points and $D$ is the number of observed dimensions. $\T$ is a $K \times 
D$ matrix with rows $\bb{\theta}_k$. $\V$ is an $N \times K$ matrix with 
rows $\mathbf{v}_n$, which are $K$-dimensional vectors, where $K$ is the number 
of latent factors.
\\ \\
The $K$ latent variables for each data point are generally assumed to be 
independent a priori: $ \mathbf{v}_n \sim \prod_{k = 1}^{K} 
\mathcal{S}(v_{nk}|\bb \varphi)$, where $\mathcal{S}$ is the prior on each 
variable with hyperparameters $\bb \varphi$ (figure \ref{fig:graphMod}). The 
prior distribution  $\mathcal{S}(v_{nk})$ can be of any type. If the 
exponential family is 
Gaussian and we use Gaussian latent variables, we recover factor analysis; 
general exponential families corresponds to the well known exponential 
family  PCA models (EPCA) \cite{collins02, mohamed09}. Considering 
non-Gaussian latent variables instantiates models such as ICA or the relevance 
vector machine (RVM) \cite{levin09BlindDecon, wipf08ard} 
\\ \\
Unsupervised models with sparsity are obtained by employing sparsity-favouring 
distributions. A sparsity-favouring distribution can be any distribution with 
high excess 
kurtosis, indicating that it is highly peaked with heavy tails, or a 
distribution with a delta-mass at zero. The set of sparsity-favouring 
distributions includes the Normal-Gamma, Normal Inverse-Gaussian, Laplace (or 
double Exponential), Exponential, or generally 
the class of scale-mixtures of Gaussian distributions \cite{polson2010}. 
Distributions that encourage sparsity fall into two classes: continuous 
sparsity-favouring or spike-and-slab distributions, which give rise to  
notions of weak and strong sparsity, respectively:
\begin{myindentpar}{3mm}
\vspace{-4mm}
\textbf{Weak sparsity.} A parameter vector $\bb{\omega}$ is considered to be 
`weakly sparse' if none of its elements are exactly zero, but has most 
elements close to zero with a few large entries. This implies that a 
weakly sparse vector $\bb{\omega}$ has a small $L_p$ norm for small $p$, or 
has entries that decay in absolute value according to some power law 
\cite{johnstone2004}. \\
\textbf{Strong sparsity.} A parameter vector $\bb{\omega}$ is considered to 
be `strongly sparse' if elements of $\bb{\omega}$ are exactly zero. The 
spike-and-slab prior places mass explicitly on zero and is thus a prior 
suited to achieving this notion of sparsity in learning.
\end{myindentpar}

%
\section{Strongly Sparse Bayesian Models} 
\label{sect:spikeslab}
A Bayesian approach to learning averages model parameters and variables 
according to their posterior probability distribution given the data, rather 
than searching for a single best parameter setting as in an optimisation 
approach. To obtain Bayesian models with strong sparsity, we use a spike-and-slab prior 
\cite{mitchellSpikeSlab,ishwaran2005}: a discrete-continuous mixture of a 
point mass at zero referred to as the `spike' and any other distribution known as the 
`slab'. 
This slab distribution is most often a uniform or Gaussian 
distribution, but may be any appropriate distribution. Since we have positive 
mass on zero, any samples produced include exact zeroes, thereby 
enforcing strong sparsity.
The spike-and-slab can also 
be seen as placing a penalty on the number of non-zero parameters, and thus enforces 
sparsity in a manner similar to an $L_0$ norm penalisation. MCMC allows us to 
stochastically find suitable solutions in this setting, where this is 
not possible otherwise due to the combinatorial nature of the optimisation.
\\ \\
We construct a spike-and-slab prior using a binary indicator matrix 
$\mathbf{Z}$ to indicate whether a latent dimension contributes to explaining 
the observed data or not. Each observed data point $\mathbf{x}_n$ has a 
corresponding vector of Bernoulli indicator variables $\mathbf{z}_n$. The spike components are combined with a Gaussian distribution, which forms 
the slab component:
\vspace{-1.4mm}
\begin{align}
& p(\mathbf{z}_n | \bb{\pi}) \! =\! \prod_k \mathcal{B}(z_{nk} | \pi_k) \!= \! 
\prod_k {\pi_{k}}^{z_{nk}} (1\!- \!\pi_{k})^{1\! -\! z_{nk}}; \\ 
& p(\mathbf{v}_n | \mathbf{z}_n, \bb{\mu}, \bb{\Sigma}) = {\prod}_k 
\mathcal{N}(v_{nk} | z_{nk} \mu_k, z_{nk} \sigma_k^2), \label{eq:slabPart} 
\end{align}
where $\mathcal{N}$ represents the Gaussian density with mean 
$\mu_k$ and variance $\sigma_k^2$. We place 
a Beta prior $\beta(\pi_k | e, f)$ on the Bernoulli parameters $\pi_k$.  When $z_{nk} = 0$ , 
$p(v_{nk})$ in equation \ref{eq:slabPart} becomes a $\delta$-function at zero, 
indicating the spike being chosen instead of the slab. We complete the 
model specification by using a Gaussian-Gamma prior for the unknown mean $\mu_k$ and 
variance $\sigma_k^2$ . We denote the set of unknown variables to 
be inferred as $\bb{\Omega} = \{ \mathbf{Z}, \mathbf{V}, 
\bb{\Theta}, \bb{\pi},  \bb\mu, \bb\Sigma \}$ and the set of 
hyperparameters $\bb{\Psi} = \{e, f, \bb{\lambda}, \nu \}$.
\\  \\
\textbf{MCMC Sampling Scheme} \\
Since the spike-and-slab is not differentiable, many popular MCMC techniques, such as Hybrid Monte Carlo, are not applicable.
We 
proceed in the context of Metropolis-within-Gibbs sampling, where we 
sequentially sample each of the unknown variables using Metropolis-Hastings. 
Our sampling procedure iterates through the following steps : 1) Sample 
$\mathbf{Z}$ and $\mathbf{V}$ jointly; 2) Sample $\bb\Theta$ by slice sampling 
\cite{nealSlice2003}; 3) Sample $\bb\mu$, $\bb\Sigma$ and $\bb\pi$ by Gibbs 
sampling.
\\ \\
In sampling the latent factors $z_{nk}$ and $v_{nk}$ in step 1, we first 
decide whether a latent factor contributes to the data or not by sampling 
$z_{nk}$ having integrated out $v_{nk}$: $p(z_{nk}\!=\! 0 | \mathbf{X}, \bb 
\pi,\! \mathbf{V}_{\neg nk})$ and $p(z_{nk}\! =\! 1 | \bb X, \bb \pi, 
\!\mathbf{V}_{\neg nk})$, where $\mathbf{V}_{\neg nk}$ are current values of  
$\mathbf{V}$, with $v_{nk}$ excluded. Based on this decision, the latent 
variable is sampled from the spike or the slab component. All variables 
$v_{nk}$ associated with the slab components are sampled using slice sampling. 
Evaluating these probabilities  involves computing the 
following integrals:
\begin{align}
p(z_{nk}\!\! & =\!\! 0 | \mathbf{X},\! \bb \pi,\! \mathbf{V}_{\!\neg nk}) \! = 
\!\!\! \int \!\! p(z_{nk}\!\! = \!0, v_{nk}\!\!  =\! 0 | \mathbf{X},\!\! \mathbf{V}_{\!\neg nk}, 
\!\bb\pi )d v_{nk}\!\! \nonumber \\
 & =\! (1\! -\! \pi_k)  p(\mathbf{X} | \mathbf{V}_{\neg nk},\! v_{nk}\!\! =\!0,\! \bb\Theta) 
\label{eq:z0Int}
\end{align}
\begin{align}
\vspace{-3mm}
\! p(z_{nk}\!\! & =\! 1 | \mathbf{X},\!\! \mathbf{V}_{\neg nk})\!  = 
\!\!\! \int \!\! p(z_{nk}\!\! =\! 1,\! v_{nk} | \mathbf{X},\!\!\mathbf{V}_{\neg nk},\! \bb\pi ) d v_{nk} \nonumber \\
& = \pi_k\!\! \int \!\! p(\mathbf{X} | \mathbf{V}, \bb\Theta) 
\mathcal{N}(v_{nk}|\mu_k, \sigma_k^2) d v_{nk} \label{eq:z1Int}
\end{align}
While computing equation \ref{eq:z0Int} is easy, the integral in equation 
\ref{eq:z1Int}  is not tractable in general.  In the case of the Gaussian family, $v_{nk}$ can be marginalised and we do exactly this. 
For other families the integral must be approximated. A number of 
approximation methods exist such as Monte Carlo integration, importance 
sampling and pseudo-marginal approaches, and the Laplace 
approximation, which we use here. The use of Laplace's method introduces a 
bias due to the approximation of the target distribution. This problem has 
been studied by \citet{rousseau2005} where the Laplace approximation is used 
in MCMC schemes with latent variables such as in our case, and show that such 
an approach can behave well. \citet{rousseau2005} show that as the number of 
observations increases, the approximate distribution becomes close to the true 
distribution, and describe a number of assumptions for this to hold, such as 
requiring differentiability, a positive definite information matrix and 
conditions on the behaviour of the prior at boundaries of the parameter space.
\\ \\
At least three other approaches for sampling  the latent variables can be considered: 1) A more naive sampling of alternating between $\mathbf{V}$ and $\mathbf{Z}$ without integrating out the slab. 2) Sampling $\mathbf{V}$ after integrating $\mathbf{Z}$. We found the collapsed scheme we describe in eq \eqref{eq:z0Int}--\eqref{eq:z1Int} quickly informs us of the state of the slab overall and resulted in faster mixing. 3) Reversible jump MCMC is also feasible and requires a different prior specification, also using a binary indicator vector but with a prior on the number of non-zero latent variables (e.g., using a Poisson). 
\\ \\
We sample $\mathbf{V}$ and $\bb\Theta$ in steps 1 and 2 by slice sampling 
\cite{nealSlice2003}, which can be thought of as a general version of the 
Gibbs 
sampler. Sampling proceeds by alternately sampling an auxiliary variable $u$, 
the slice level, and then 
randomly drawing a value for the parameter from an interval along the slice. 
The variables \{$\bb\mu$, $\bb\Sigma$\} and $\bb\pi$ in step 3 have conjugate 
relationships with the latent variables $\mathbf{V}$ and $\mathbf{Z}$ 
respectively. Gibbs sampling is used since the full conditional distributions are easily derived\footnote{Implementation notes online at: \texttt{cs.ubc.ca/\textasciitilde shakirm}}. 

\section{Models with $L_1$ norms and Sparsity-Favouring Priors}
\label{sect:l1sparsity}
The $L_1$ norm has become 
the established mechanism with which to encode sparsity into many problems, and has a strong connection to continuous densities that promote sparsity.
The $L_1$ norm has a number of appealing properties: it gives the closest 
convex optimisation problem to the $L_0$ problem; there is an broad 
theoretical basis with provable properties ($L_0\!\!-\!\!L_1$ equivalence and 
exact recovery based on RIP); and can be implemented efficiently based on the tools 
of convex optimisation (linear and semi-definite programming).  
\\ \\
\textbf{Sparsity Inducing Loss Functions} \\
This leads us naturally to consider sparse latent variable models based on the
$L_1$ norm. If we assume that the latent distribution 
is a Laplace, $\mathcal{S}(\mathbf{v}_n) \propto \exp(-\alpha \| \mathbf{v}_n 
\|_1)$, the maximum a posteriori solution  for $\V$ is equivalent to $L_1$ 
norm regularisation in this model. We define the following objective for 
sparse generalised latent variable modelling: 
\begin{equation}
\vspace{-1.5mm}
\label{eq:mapLoss}
\min_{V, \Theta} {\sum}_n \ell\left(\mathbf{x}_n, \mathbf{v}_n \bb{\Theta} 
\right) + \alpha \|\mathbf{V}\|_1 + \beta R(\bb\Theta),
\end{equation}
where the loss function $ \ell\left(\mathbf{x}_n, \mathbf{v}_n \bb{\Theta} 
\right)\!\!=\!\!-\ln p(\mathbf{x}_n | \mathbf{v}_n\bb{\Theta})$, is 
the negative log likelihood obtained using equation \ref{eq:exponSpec}. 
Equation \ref{eq:mapLoss} provides a unifying framework for sparse models with 
$L_1$ regularisation.  The regularisation parameters $\alpha$ and $\beta$, 
control the sparsity of the latent variables and the degree to which 
parameters will be penalised during learning. The function $R(\bb \Theta)$ is 
the regulariser for the model parameters $\bb\Theta$. This model is specified 
generally and applicable for a wide choice of regularisation functions 
$R(\cdot)$, including the $L_1$ norm. Such a loss function was 
described previously by \citet{lee09sepca} -- here we focus on unsupervised 
settings and specify the loss more generally, allowing for both sparse 
activations as well as basis functions. One configuration we consider is the 
use of the modified loss \eqref{eq:mapLoss} with $R(\bb{\Theta})\!\!=\!\!-\ln 
p(\bb{\Theta} | \bb{\lambda}, \nu) $. This loss allows sparsity in the 
latent variables and corresponds to finding the maximum a posteriori (MAP) 
solution. We shall refer to this model as the $L_1$ model.
\\ \\
Optimisation is performed by alternating minimisation. Each 
step then reduces to established problems for which, we can then rely on the 
extensive literature regarding $L_1$ norm minimisation. A number of 
methods exist to solve these problems: they can be recast as 
equivalent inequality constrained optimisation problems and solved using a 
modified LARS algorithm \cite{leeL1reg2006}, recast as a second order cone 
program, or solved using a number of smooth approximations to the 
regularisation term \cite{schmidt2007}, amongst others. 
\\ \\
\textbf{Sparse Bayesian Learning}\\
Continuous densities with high excess kurtosis such as the zero-mean 
Laplace distribution or Student's-$t$ distribution are often used in Bayesian 
models where sparsity is desired. For a model with priors that prefer sparsity, the Bayesian averaging process often results in non-sparse posteriors and give solutions that are nearly zero, resulting in weakly sparse models.  We consider two models with sparsity in the latent variables $\mathbf{v}_n$:
\vspace{-3mm}
\begin{myindentpar}{3mm}
\textbf{Laplace Model.} Using the Laplace distribution: $\mathbf{v}_n 
\sim \prod_{k = 1}^{K} \tfrac{1}{2}b_k \exp \left( - b_k|v_{nk}| \right)$, a 
Bayesian version of the $L_1$ model described by equation \ref{eq:mapLoss} 
can be specified. The equivalence between this model and the $L_1$ model can 
be seen by comparing the log-joint probability using the Laplace distribution, 
to the $L_1$ loss of equation \ref{eq:mapLoss}. We refer to
Bayesian inference in this Laplace model as LXPCA, in contrast to the $L_1$ 
model, which is an optimisation-based method.
\\
\textbf{Exponential Model.} If parameters or latent variables are to be positively 
constrained, the natural choice would be an exponential distribution peaked at 
zero: $\mathbf{v}_n \sim 
\prod_{k = 1}^{K} b_k \exp \left( - b_k v_{nk} \right)$, which has similar shrinkage 
properties to the Laplace.  We refer to this model as NXPCA.
\end{myindentpar}
\vspace{-1mm}
These distributions are popular in sparse regression problems \cite{seeger07, 
wipf08ard} and are natural candidates in the unsupervised models explored 
here. The hierarchical model 
specification is completed by placing a Gamma prior on the unknown rate 
parameters $\mathbf{b}$, with shared shape and scale parameters $\alpha$ and 
$\beta$ respectively. We denote the set of unknown variables to be inferred as 
$\bb{\Omega} = \{ \V, \T, \mathbf{b} \}$ and the set of hyperparameters 
$\bb{\Psi} = \{ \alpha, \beta, \bb{\lambda}, \nu \}$. 
The joint probability of the model is:
\begin{flushright}
\begin{equation}
\label{eq:jointProb}
\! p(\mathbf{X},\bb{\Omega} | \bb{\Psi}) \! = \! p(\mathbf{X} | 
\mathbf{V}, \bb{\Theta}) p(\bb{\Theta} | \bb{\lambda}, \nu) p(\mathbf{V} | 
\mathbf{b}) p(\mathbf{b} | \alpha, \beta)
\end{equation}
\end{flushright}
Inference in this model is accomplished using Markov Chain Monte Carlo (MCMC) 
methods, and the log of the joint probability \eqref{eq:jointProb} is central 
to 
this sampling. We use a sampling approach based on Hybrid Monte Carlo 
(HMC). This can be implemented easily, and we defer the 
algorithmic details to 
\citet{book:mackay}. 
\vspace{-1mm}
\section{Related Work}
The body of related work is broad and the work described here is far from 
exhaustive, but attempts to capture many papers of relevance in 
contextualising approaches to, and applications of sparse learning.
There is a wide body of literature for sparse learning in problems of feature 
selection, compressed sensing and regression using the 
$L_1$ norm, such as those by \citet{lasso96, aspremont05, candes06, leeL1reg2006}. 
Bayesian methods for sparse regression problems using 
continuous distributions have also been discussed by \citet{, seeger07, carvalho2008horse, ohara2009}. 
\citet{wipf08ard} derive a relationship between 
automatic relevance determination (ARD), maximum likelihood and iterative 
$L_1$ optimization. \citet{archambeau09probproj} provide a nice 
exploration of ARD-related priors and variational EM for sparse PCA and sparse CCA.
\\ \\
Of relevance to unsupervised learning of real-valued data is sparse PCA and 
its variants \cite{zou04sparse, aspremont05, Rattray2009}.
The wide body of 
literature on matrix factorisation is also indirectly related 
\cite{Airoldi08mixedmembership}.
These methods do not deal with the 
exponential family generalisation and may yield sparse factors as a 
by-product, rather than by construction. There are also many other papers of 
relevance in bioinformatics, computer vision, ICA and blind deconvolution 
\cite{levin09BlindDecon}. The methods we develop here also have a strong bearing on the 
basis pursuit problem widely used in geophysics and other engineering fields and can allow not 
only for the solution of basis pursuit, but also in obtaining useful estimates of uncertainty.
\\ \\
The use of `spike-and-slab' sparsity for 
variable selection was established in statistics by \citet{mitchellSpikeSlab} 
and more recently by \citet{ishwaran2005}. \citet{yen2011majorization} 
describes a majorisation--minimisation algorithm for MAP estimation,
and \citet{lucke2012} describe EM for Gaussian sparse coding.
\citet{carvalho2008spike} use spike-and-slab-type priors to 
introduce sparsity in Bayesian factor regression models. They consider a 
hierarchical sparsity prior to reduce uncertainty as to whether a parameter is 
non-zero. This comes with increased computation and may not necessarily 
improve performance. \citet{courville2010ssrbm} describe spike-and-slab 
for deep belief networks. 
\section{Experimental Results}
\label{sect:experiments}
We consider the generalisation performance of unsupervised 
methods to unseen data, which appear as missing data. To handle missing data, 
we divide the data into a set of observed and missing data, $\mathbf{X} = 
\{\mathbf{X}^{obs},  \mathbf{X}^{missing}\}$ and condition on the set 
$\mathbf{X}^{obs}$ in the inference. 
We create test sets by randomly selecting 10\% of the elements of the 
data matrix. Test elements are set as \textit{missing values} in the 
training data, and our learning algorithms have been designed in all cases to 
handle missing data. We calculate the predictive probability (negative log 
probability, NLP) and the root mean squared error (RMSE) using the testing 
data. We created 20 such data sets, each with a different set of missing data, 
and provide mean and one standard deviation error bars for each of our 
evaluation metrics. 
For fairness, 
the regularisation parameters $\alpha$ and 
$\beta$ in section \ref{sect:l1sparsity} are chosen by cross-validation using 
a validation data set, which is chosen as 5\% of the data elements. This set 
is independent of the data that has been set aside as training or testing 
data. 
\vspace{-2mm}
\subsection{Benchmark Data}
We use the block images data \cite{ibp} as a synthetic benchmark data set.
The data consists of binary images, with each image 
$\mathbf{x}_n$ represented as a 36-dimensional vector. The images were 
generated with four latent features, each being a type of block. The 
observed data is a combination of a number of these latent features. 
Noise is added by flipping bits in the images with probability 0.1.
This data set consists of a number of latent factors, 
only a subset of which contributes
to explaining any single data point.  This data is synthetic, but not generated from any of the models tested.
\begin{figure*}[t]
\centering
\subfigure[]{
\includegraphics[height =3.7cm]{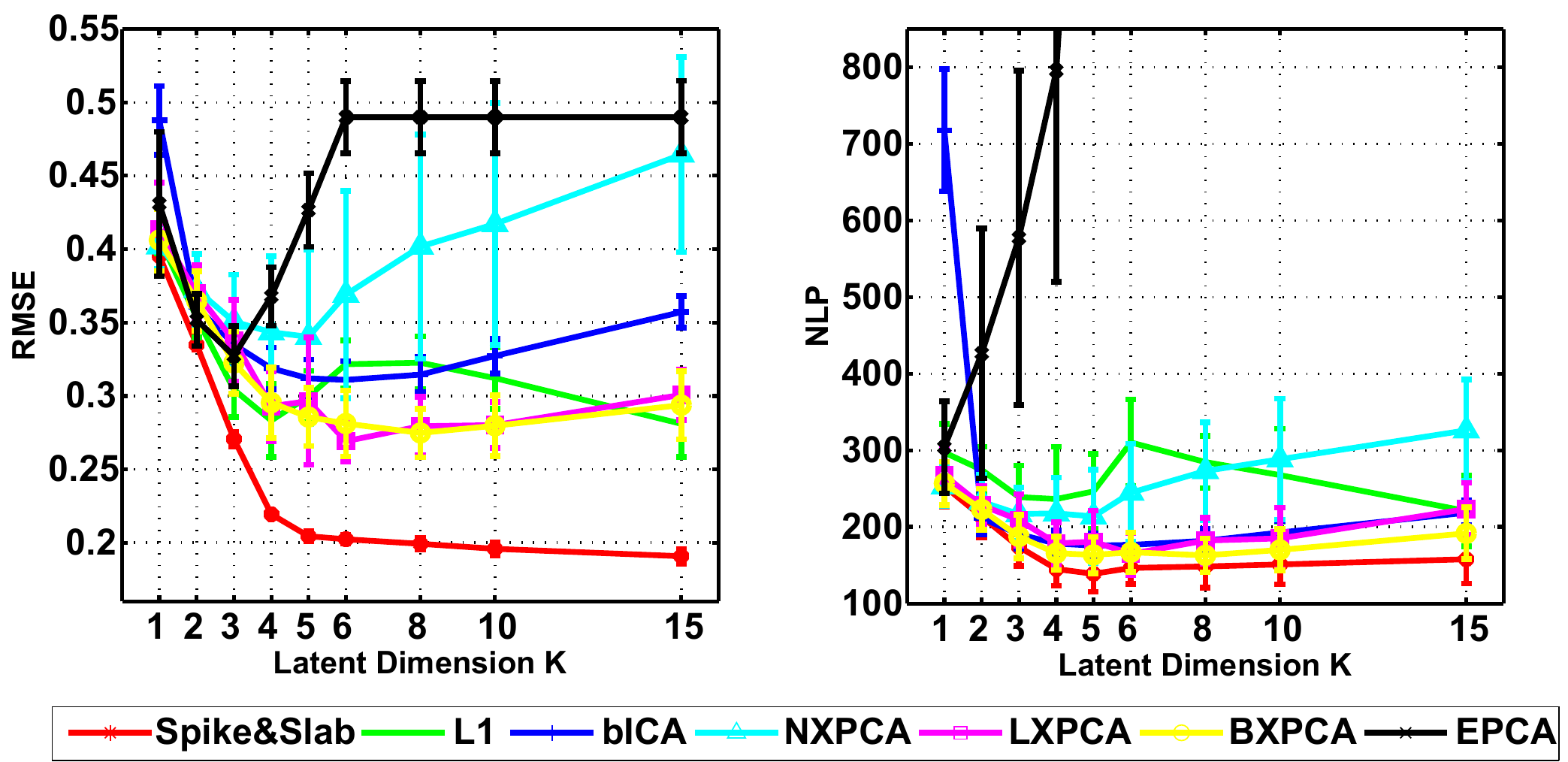}
\label{fig:synthDataPlot}
}
\subfigure[]
{
	\includegraphics[height = 3.7cm]{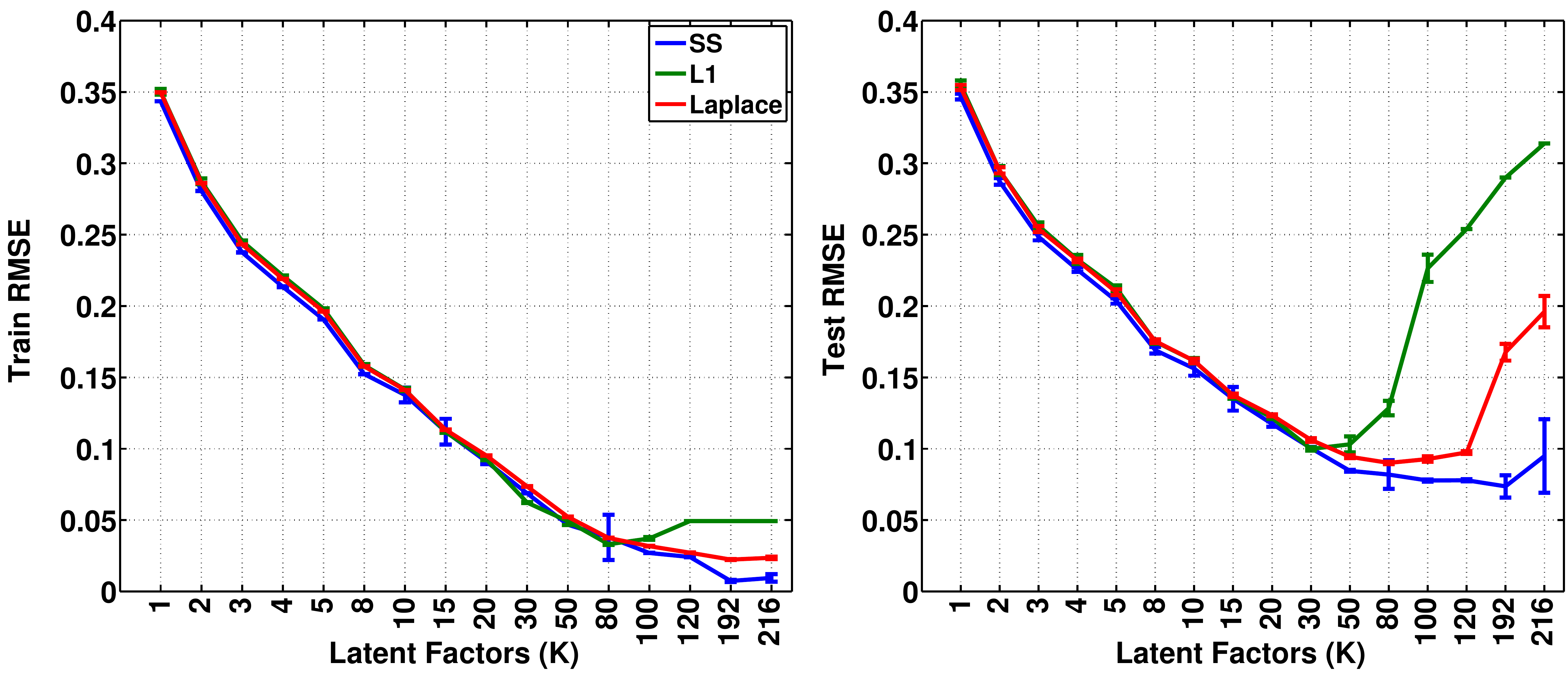}
	\label{fig:natScenes}
}
\vspace{-4mm}
\caption{\small (a) RMSE and NLP for various latent dimensions on the block 
images data set (binary). 
	(b) Performance in terms of RMSE on natural scenes (real-valued). 
	}
\end{figure*}
\\ \\ \noindent
Figure \ref{fig:synthDataPlot} shows the NLP and RMSE on this benchmark data set. The methods developed are compared to EPCA \cite{collins02}, BXPCA \cite{mohamed09} and to binary ICA \cite{kaban06}. A random predictor would have an $NLP\!=\! 100\! \times\! 36\! \times\! 10\%\! =\! 360$ bits. The models tested here have performance significantly better than this. 
Both optimisation-based and Bayesian learning approaches do well, but
the spike-and-slab model shows the best performance with smaller error bars. 
\vspace{-2mm}
\subsection{Real Data}
\begin{figure}[t]
\centering
\includegraphics[height = 3.5cm]{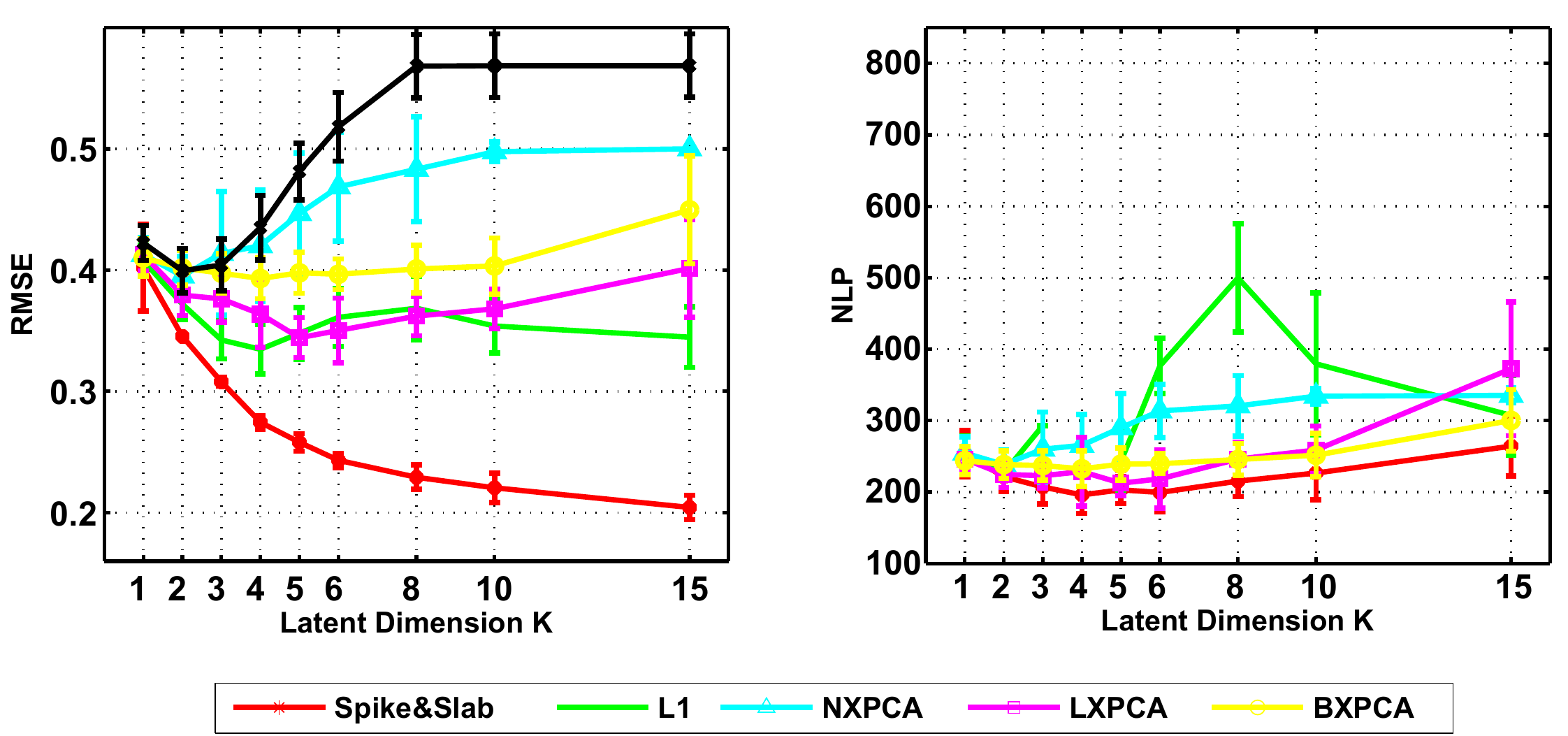}
\caption{\small RMSE and NLP for the aimal attributes data.}
\label{fig:animalResPlot}
\vspace{-8mm}
\end{figure}

We summarise the real data sets we use in table \ref{tab:realData} (which includes data in the $D>N$ regime).
\begin{table}[t]
\scriptsize
\centering
\caption{Summary of real data used.}
\label{tab:realData}
\begin{tabular}{|c|l|c|c|l|}
\hline
\# & Data & N & D & Type\\
\hline \hline
1 & Natural scenes & 10,000 & 144 & Real\\
2 & Animal attributes & 33 & 102 & Binary\\
3 & Newsgroups & 100 & 200 & Counts\\
4 & Hapmap & 100 & 200 & Binary\\
\hline
\end{tabular}
\vspace{-3mm}
\end{table}
\textbf{Natural images} are the topic of much research based on $L_1$ regularisation. For the \citet{olshausen1996} image data set,
we use $12 \times 12$ image patches extracted from a set of larger images. We 
use the \textit{Gaussian} instantiation of the sparse generalised model (equation 
\ref{eq:exponSpec}) and evaluate the performance of: 
$L_1$ optimisation; a Laplace-prior factor model; and 
the Bayesian spike-and-slab model. Our results are shown for both 
underdetermined and overcomplete bases (K = 192 as in \citet{olshausen1996}) in 
figure \ref{fig:natScenes}. All methods perform similarly in the low-rank 
approximation cases, but as the model becomes overcomplete, Bayesian methods 
perform better with the spike-and-slab method much better than other methods, particularly
in reconstructing held-out/missing data. 
The \textbf{animal attributes} data set of \citet{kemp2008} 
consists of animal species with ecological and biological properties as 
features. We use the \textit{binary unsupervised model} and show 
results for various latent dimensions for NLP and RMSE in figure 
\ref{fig:animalResPlot}. For this data, the NLP of a random classifier is 336 
bits and the models have NLP values much lower than this.
\begin{figure*}[t]
		\centering
		\subfigure[Animal Attr.]{
			\includegraphics[height = 2.8cm]{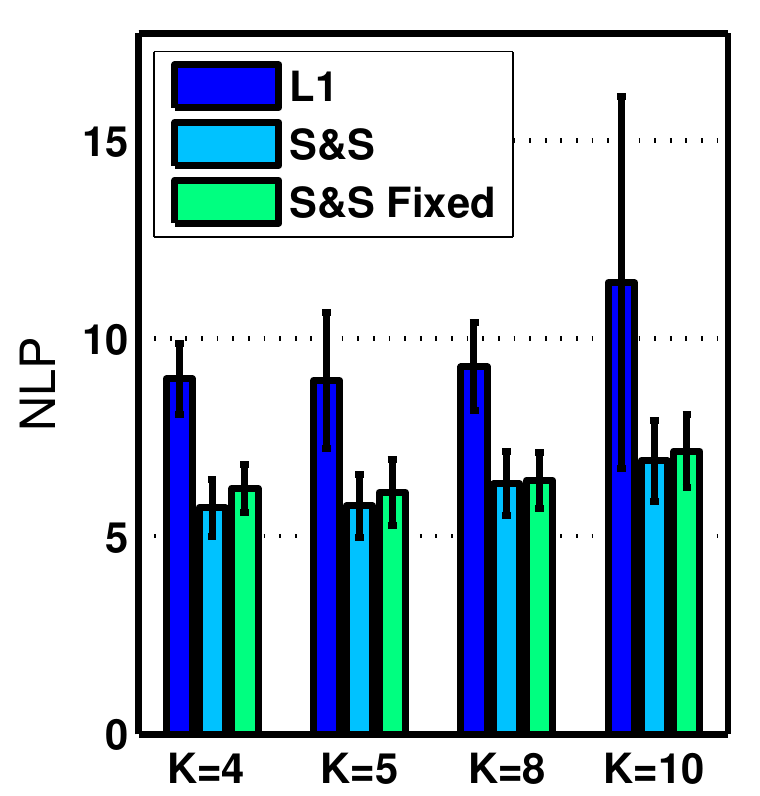}
			\label{fig:animTimeNLP}
		}
		\subfigure[Newsgroups]{
			\includegraphics[height = 2.8cm]{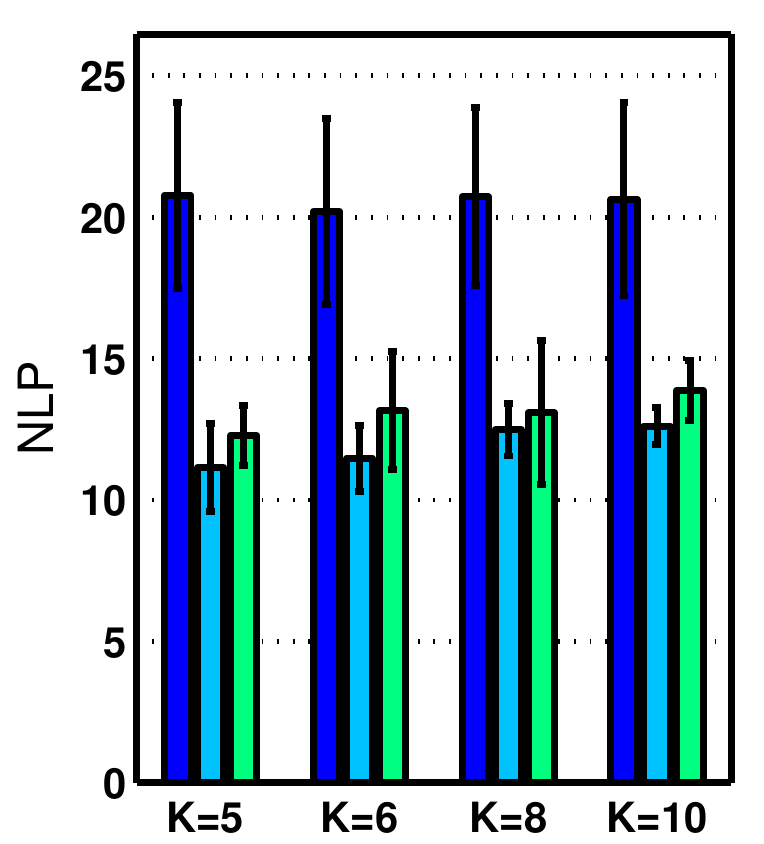}
			\label{fig:newTimeNLP}
		}
		\subfigure[Hapmap Data]{
				\includegraphics[height = 2.8cm]{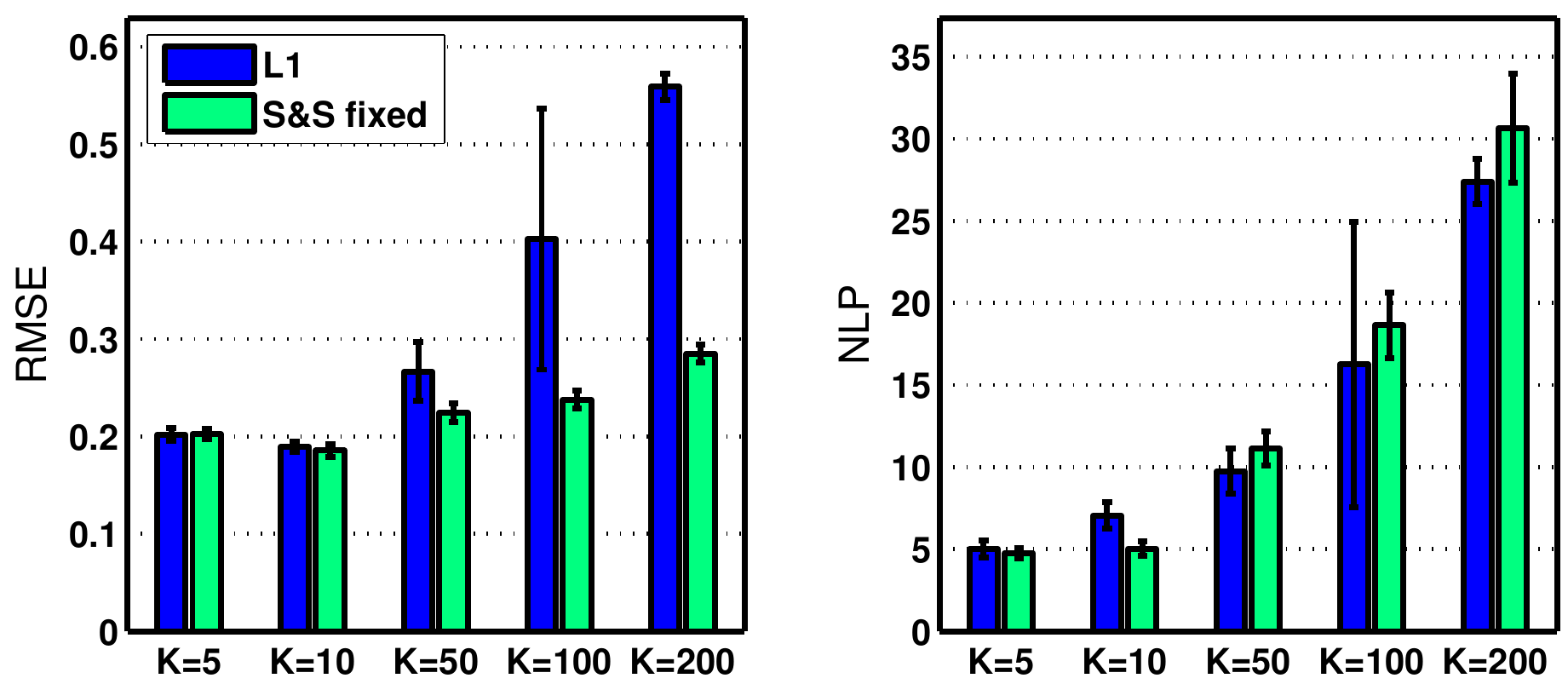}
				\label{fig:hapmapRes}
		}
		\subfigure[Newsgroups sparsity]{
		\begin{minipage}{0.23\linewidth}
		\vspace{-20mm}
		\scriptsize
				\begin{tabular}{|p{2mm}|c|c|}
					\hline
					\textbf{K}& \textbf{$L_1$} & \textbf{Spike-Slab}\\
					\hline
					5& 475 $\pm$36  & 1446 $\pm$24\\
					\hline
					6& 483$\pm$57 & 1418$\pm$ 29\\
					\hline
					8& 592$\pm$207 & 1400$\pm$18\\
					\hline
					10& 934$\pm$440 & 1367$\pm$32\\
					\hline
				\end{tabular}
				\label{tab:NNZ}
			\vspace{5mm}	
		\end{minipage}
		}
		\vspace{-5mm}
		\caption{\small (a) - (c) Comparison of predictive probabilities (NLP). `S\&S fixed' is the time-matched 
spike-and-slab performance (elaborated upon in sect. \ref{sect:concl}). 
(d) Num. of non-zeros in newsgroups reconstruction - the true number is 1436.
}
		\label{fig:timeNLP}
		\vspace{-3mm}
\end{figure*}
\\ \\ \noindent
We also use a subset of the popular \textbf{20 newsgroups} data set, consisting of documents and counts of the words used in each document, with data sparsity of 93\%. 
Figure \ref{fig:newTimeNLP} shows the performance of the \textit{Poisson unsupervised model}
using $L_1$ and spike-and-slab. Apart from the application of the model to 
count data, the results show that the spike-and-slab model is able to deal 
effectively with the sparse data and provides effective reconstructions and 
good predictive performance on held out data. 
We are also able to show the improved behaviour of the spike-and-slab model using the \textbf{Hapmap} data set\footnote{Obtained from: https://mathgen.stats.ox.ac.uk/impute/}. 
The comparative performance is shown in figure \ref{fig:hapmapRes} showing the 
spike-and-slab has performance similar to $L_1$ in terms of RMSE at low $K$, 
but much better performance for large $K$. 
\section{Discussion and Conclusion} \normalsize
\label{sect:concl}
The common lore when using MCMC is that it is dramatically slower than 
optimisation methods. 
For optimisation methods, the cross-validation procedure 
needed to set regularisation parameters $\alpha$ and $\beta$, is 
computationally demanding due to the need to execute the optimisation for many 
combinations of parameters. This approach is also wasteful of data, since a 
separate validation data set is needed to make sensible choices of these 
values and to avoid model overfitting. While individual optimisations may be 
quick, the overall procedure can take an extended time, which depends on the 
granularity of the grid over which regularisation values are searched for. 
These parameters can be learnt in the Bayesian setting
and  have the advantage that we obtain information about the distribution of our latent variables, 
rather than point estimates and can have significantly better performance. 

Figure \ref{fig:timeNLP} demonstrates this tradeoff between running time and 
performance of the optimisation and the Bayesian approaches. $L_1$ was allowed 
to run to convergence and the spike-and-slab for 200 iterations. In this instance, the Bayesian 
method is seemingly slower, but produced significantly better reconstructions in both the 
human judgements and newsgroups data.
We considered the 
setting where we have a fixed time budget and
fixed the running time for the spike-and-slab to that used by the $L_1$ model (including time to search for hyperparameters). The 
result is shown (as S\&S fixed) in figure \ref{fig:timeNLP}, which shows that 
even with a fixed time budget, MCMC performs better in this setting. The table of figure 
\ref{tab:NNZ} shows that the number of non-zeroes in the reconstructions for 
various $K$ for the newsgroups data, with the true number of non-zeroes being 
1436. $L_1$ is poor in learning the structure of this sparse data set, whereas 
the spike-and-slab is robust to the data sparsity. 

All our results showed the spike-and-slab approach to have better performance 
than other methods compared in the same model class. The models based on the 
$L_1$ norm or Bayesian models with continuous sparsity favouring priors 
enforce global shrinkage on parameters of the model. It is this property that 
induces the sparsity property, but which also results in the shrinkage of 
parameters of relevance to the data. This can be problematic in certain cases, 
such as the newsgroups data set which resulted in overly sparse data 
reconstructions.
The spike-and-slab has the ability to give both global and local 
shrinkage, thus allowing sparsity in the model parameters while not 
restricting parameters that contribute to explaining the data. 

Current approaches for sparse learning will have difficulty scaling to large data sets in this regime. We might think of EP as a potential solution, such as used by \citet{hernandez2010expectation}, but this is restricted to regression problems.
For the standard Gaussian model, \citet{Rattray2009} discuss this issue and propose a hybrid VB-EP approach as one way of achieving fast inference, but such an approach is not ideal, 
leaving scope for future work. 

We have 
demonstrated that improved performance can be obtained by considering sparse 
Bayesian approaches. In particular, Bayesian learning with spike-and-slab 
priors consistently showed the best performance on held out data and produced 
accurate reconstructions, even in the `large $p$' paradigm or with restricted 
running times. By considering the broad family of unsupervised latent variable 
models, we developed a sparse generalised model and provided new sampling 
methods for sparse Bayesian learning using the spike-and-slab distribution. 
Importantly, we have provided the first comparison of sparse unsupervised
learning using three approaches: optimisation using the $L_1$ norm, Bayesian 
learning using continuous sparsity favouring priors, and Bayesian learning 
using the spike-and-slab prior. We have also demonstrated our methods in 
diverse applications including text modelling, image coding and psychology 
showing the flexibility of the sparse models developed. These results show 
that Bayesian sparsity and spike-and-slab methods warrant a more prominent 
role and wider use in sparse modelling applications.

\begin{small}
\textbf{Acknowledgements} \\ 
Support for: SM by the Canadian Institute for Advanced Research (CIFAR); KAH by an NSF Postdoctoral Fellowship; ZG by EPSRC EP/I036575/1.
\end{small}
\vspace{-3mm}
\bibliographystyle{plainnat}
\bibliography{refs}

\end{document}